# Review on Parameter Estimation in HMRF

*Georgia Institute of Technology*

*Namjoon Suh*

*Abstract*

*This is a technical report which explores the estimation methodologies on hyper-parameters in Markov Random Field and Gaussian Hidden Markov Random Field. In first section, we briefly investigate a theoretical framework on Metropolis-Hastings algorithm. Next, by using MH algorithm, we simulate the data from Ising model, and study on how hyperparameter estimation in Ising model is enabled through MCMC algorithm using pseudo-likelihood approximation. Following section deals with an issue on parameters estimation process of Gaussian Hidden Markov Random Field using MAP estimation and EM algorithm, and also discusses problems, found through several experiments. In following section, we expand this idea on estimating parameters in Gaussian Hidden Markov Spatial-Temporal Random Field, and display results on two performed experiments.*

**1. Review on Metropolis – Hastings algorithm**

*Metropolis-Hastings algorithm* is a Markov Chain Monte Carlo method for obtaining a sequence of random samples from a probability density for which direct sampling is difficult. In other words, constructing a Markov Chain which converges to the target distribution $\pi(x)$ is enabled by Metropolis-Hastings algorithm. Their basic idea is as follows:

In order for the Markov Chain $\{x^0, x^1, x^2, x^3 \ldots\}$ to converge to the target density, $\pi(x)$, we should be able to know how to construct a transitional kernel, whose nth (n→ ∞) iteration converge to the invariant target distribution.

Let a $q(y|x)$ be a candidate-generating density (*or proposal density*). This density is to be interpreted as saying that when a process is at the point $x$, the density generates a new candidate data y. If $q(y|x)$, itself, satisfies "Detailed balance condition" for all the x, y $\in \chi$, $x \neq y$. ($\chi$ denotes a continuous state space, where the Markov Chain is defined), then our search is over. But in most cases, it won't.

$$\pi(x)q(y|x) > \pi(y)q(x|y) \tag{1}$$

In above case, speaking somewhat loosely, the process moves from $x$ to $y$ too often, and from $y$ to $x$ rarely. Metropolis and Hastings introduced an "acceptance probability" allowing the Markov Chain to move towards the direction where the process can satisfy the "Detailed Balance condition", so that the data x and y can be generated from target distribution, $\pi(.)$. Following is the result of introducing the "acceptance probability", satisfying the detailed balance equation.

$$\pi(x)q(y|x)\alpha(x,y) = \pi(y)q(x|y)\alpha(y,x) \qquad (2)$$

Here $\alpha(x,y)$ denotes that we accept the movement from x to y with probability $\alpha(x,y)$. The above inequality, (1), tells us that movement from y to x is not made enough. We should therefore define $\alpha(y,x)$ as large as possible. Since it is a probability, its upper limit is 1. If we plug in 1 at $\alpha(y,x)$ in (2), the $\alpha(x,y)$ can be defined as follows:

$$\alpha(x,y) = \min\{1, \frac{\pi(y)q(x|y)}{\pi(x)q(y|x)}\}$$

In this case, we generate a random number from $U(0,1)$, then if the acceptance probability $\alpha(x,y)$ is bigger than the newly generated random number, then we allow the movement from x to y, otherwise, point x would stay at the same point x. Here, we only consider the case of $x \neq y$, therefore the transition kernel can be defined as follows:

$$K(x \mapsto y) = q(y|x)\alpha(x,y) \quad \text{if } x \neq y$$

To complete the definition of the Metropolis-Hastings chain, we must consider the possibly non-zero probability that the process remains at the x. In order to explain this concept more clearly, we would like to redefine the transitional kernel as follows:

Transition kernel, $K(x \mapsto B)$, is a conditional distribution function that represents the probability of moving $x$ to a point in a set (or interval) B. Transition kernel of the x being transferred to the region B can be expressed as follows: (Let B = dy).

$$K(x \mapsto B) = \begin{cases} q(B|x)\alpha(x,B), & x \notin B \\ 1 - \int_{x' \in \chi \backslash B} q(x'|x)\alpha(x,x'), & x \in B \end{cases}$$

Here, a candidate generating density (*or proposal density*) is denoted $as\ q(B\ |\ x)$. This density is saying that when a process is at the point $x$, the density generates a candidate data in the region B. $\alpha(x,B)$ is a probability of accepting the "proposal" of $x$ being moved to the region B. $\chi$ denotes a continuous state space, where the markov chain is defined. So, as can be seen from the above definition of $K(x \mapsto B)$, this transition instinctively can be separated in two cases. First, when the $x$ does not originally belong to the region B, we first propose a move from $x$ to a region B, then accept the proposal with probability $\alpha(x, B)$. Inversely, when $x$ is in the region B, the only thing that we need to do is to subtract the probability of newly moved point being in continuous state space $\chi \backslash B$ from 1. Above representation can be expressed neatly in one line as follows:

$$K(x \mapsto B) = q(B \mid x)\, \alpha(x, B) + \mathbb{1}(x \in B)\left\{1 - \int_{x' \in \chi} q(x' \mid x)\, \alpha(x, x')\right\} \tag{3}$$

For simplicity of notation, we will put $r(x) = 1 - \int_{x' \in \chi} q(x' \mid x)\, \alpha(x, x')$. In order for $\pi(x)$ to be an invariant distribution for transitional kernel $K(x \mapsto B)$, following detailed balance condition should be satisfied.

$$\pi(x) K(x \mapsto B) = \pi(B) K(B \mapsto x)$$

If we take the integral with respect to x on the left hand-side for whole continuous state space $\chi$, we can get following equality (3) and if we are able to show that this equality holds for the transitional kernel defined above (2), we complete the proof of Metropolis-Hastings algorithm.

$$\int_{x \in \chi} K(x \mapsto B)\, \pi(x) dx = \pi(B) \tag{4}$$

$$\int_{x \in \chi} K(x \mapsto B)\, \pi(x) dx = \int_{x \in \chi} q(B \mid x)\, \alpha(x, B) \pi(x) + \mathbb{1}(x \in B) r(x) \pi(x)\, dx$$

$$= \int_{x \in \chi} q(B \mid x)\, \alpha(x, B) \pi(x)\, dx + \int_{x \in B} r(x) \pi(x)\, dx$$

$$= \int_{x \in \chi} \int_{y \in B} q(y \mid x)\, \alpha(x, y)\, dy\, \pi(x)\, dx + \int_{x \in B} r(x) \pi(x)\, dx$$

$$= \int_{y \in B} \int_{x \in \chi} q(x \mid y)\, \alpha(y, x)\, dx\, \pi(y)\, dy + \int_{x \in B} r(x) \pi(x)\, dx$$

$$= \int_{y \in B} (1 - r(y)) \pi(y)\, dy + \int_{x \in B} r(x) \pi(x)\, dx$$

$$= \int_{x \in B} \pi(x)\, dx = \pi(B) \tag{5}$$

Note that we used a relation of (2) in the process of our derivation (5), proving that a M-H kernel has $\pi(x)$ as its invariant density. In Metropolis-Hastings algorithm, calculation of $\alpha(x, y)$ doesn't require the knowledge of normalizing constant in Ising model. And if the candidate generating density is symmetric, an important special case, $q(x \mid y) = q(y \mid x)$, then the probability of move from x to y reduces to $min\{1, \frac{\pi(y)}{\pi(x)}\}$, which will lead to the facilitations the calculation. We will use this algorithm's distinctive properties in following section for simulating the behavior of Ising model.

## 2. Generate samples from 2D Ising model via Metropolis-Hastings algorithm

Consider Binary Random Field on N x N regular lattice in $R^2$. Assume that each pixel at position (l, m) can take values either $Z_{lm} = 1$ or $Z_{lm} = -1$. This double indexing can be translated into univariate indexing by $(l, m) \mapsto i = (N-1)*l + m$. Let $Z = \{ Z_i : i \in \Omega \}$ be variables to indicate the true segmentation of the image. We assume that Z is from a finite state Markov Random Field (MRF). Here, we only consider, for simplicity, the Ising model with first order neighborhood system and no external field:

$$P(Z = z \mid \beta) = \frac{1}{\psi(Z)} \exp \left\{ \sum_{i \in \Omega} Z_i \left( \beta \sum_{j \in N(i)} Z_j \right) \right\}$$

$\psi(Z)$ denotes a partition function (or normalizing constant), which is required in to create a valid probability distribution: $\psi(Z) = \sum_{Z \in \{-1,+1\}^{|\Omega|}} \exp \left\{ \sum_{i \in \Omega} Z_i \left( \beta \sum_{j \in N(i)} Z_j \right) \right\}$. Because of this notorious partition function which is computationally infeasible to get, we are unable to simulate the behavior of Ising model only by using the ordinary sampling scheme. Here we will use Metropolis-Hastings algorithm to sample from the target density, $P(Z = z \mid \beta)$. In our model, the calculation of $\alpha(z, z')$ in Metropolis-Hastings does not require the knowledge of normalizing constant of $P(Z = z \mid \beta)$, since it appears both at the denominator and at the nominator of $\alpha(z, z')$.

Since each pair of neighboring pixels corresponds to one edge inside N x N regular grid, there is N-1 vertical edges in each of N rows, i.e. N(N-1) vertical edges. Because of the symmetry, the number of horizontal edges is the same, and there is exactly 2N(N-1) pairs. The expression $\sum_{i \in \Omega} Z_i \left( \beta \sum_{j \in N(i)} Z_j \right)$ is 2N(N-1) - $2 d_z$ where $d_z$ is the number of disagreeing edges (bordering pixels with values -1 and 1 in Z). So we can write it as follows

$$P(Z = z \mid \beta) \propto \exp(-2\beta d_z)$$

With this simplification, we can easily implement MH algorithm in our model. Start with the space of all configurations C in which each configuration Z is represented as a vector.

$$Z = (z_1 \ z_2 \ \ldots \ldots \ldots, z_i, z_{i+1}, \ldots \ldots z_{N^2})$$

The MH algorithm has a following steps:

1. Start with the random configuration Z.
2. Select a random pixel $z_i$
3. Propose a new configuration Z',
$$Z' = (z_1 \ z_2 \ \ldots \ldots \ldots, -z_i, z_{i+1}, \ldots \ldots z_{N^2})$$

4. Calculate the acceptance probability $\alpha(z, z') = \min \left\{ 1, \frac{P(Z=z'|\beta) \, g(z'|z)}{P(Z=z|\beta) \, g(z|z')} \right\}$.

   where $g(z'|z) = \begin{cases} 0 & z \text{ and } z' \, differ \, in \, more \, than \, single \, coordinate \\ \frac{1}{N^2} & z \text{ and } z' \, differ \, only \, in \, single \, coordinate \end{cases}$

5. Generate the random number between 0 and 1, $u \in U(0,1)$
   Accept newly proposed z' if $u \leq \alpha(z, z')$, otherwise, stay with original z.
6. Iterates Step 1 ~ 5 until the samples converge to the stationary state.

In Step 4, the acceptance rate can be easily computed as follows: Since proposal density is symmetry, it can be cancelled out from denominator and nominator. Then, let $d_{z_i}$ denote the number of disagreeing edges between $z_i$ and the neighborhood of $z_i$. If we flip the value of $z_i$ to $z_i' = -z_i$, we can easily notice that $d_{z_i'}$ is equivalent to $a_{z_i}$, which denotes the number of agreeing nodes between $z_i$ and its neighborhood nodes. Since $P(Z = z | \beta) \propto \exp(-2\beta d_z)$, $\frac{P(Z=z'|\beta)}{P(Z=z|\beta)}$ is equivalent to $\exp(-2\beta(a_{z_i} - d_{z_i}))$. Following pictures are the simulation results of Ising Model on 125 x 125 lattice using Metropolis-Hastings algorithm by varying the values of betas. (beta = -1, beta = 0.3, beta = 0.85). We illustrated the number of iterations for each simulation to be converged. From left to right and below, beta = -1, 0.3, 0.85 respectively.

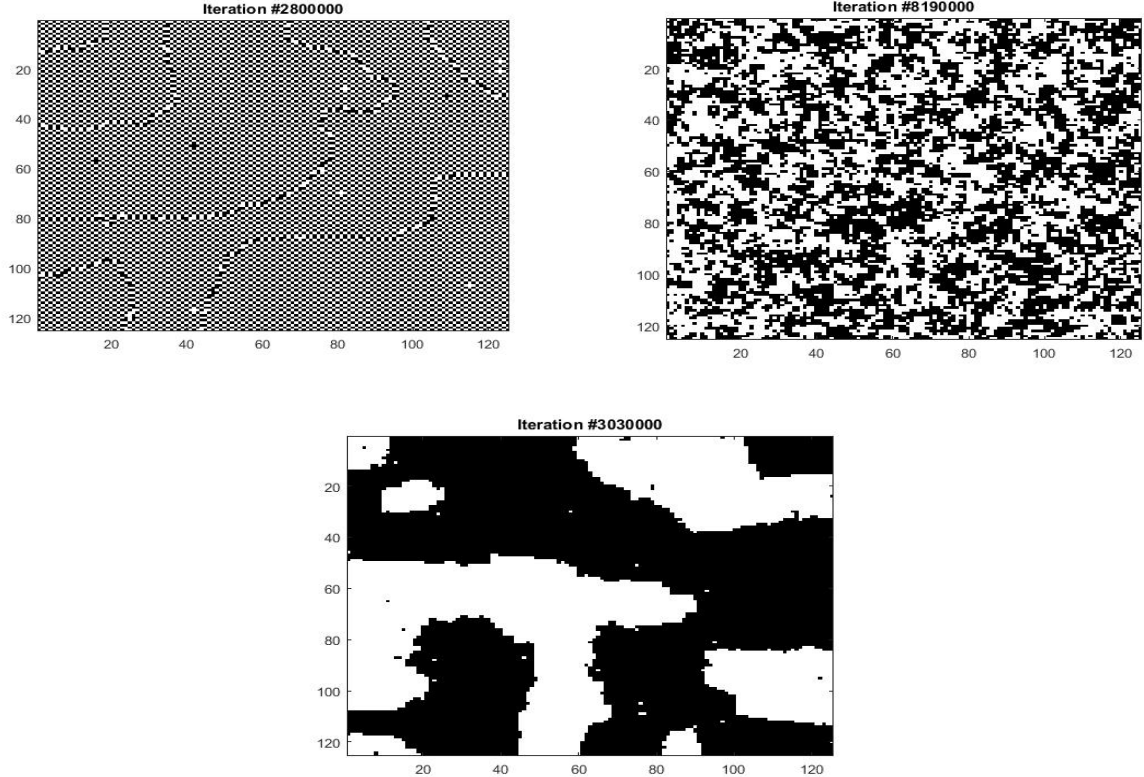

*Figure 1. Ising model with beta = -1 (Left above), 0.3 (Right above), 0.85 (Below)*

Here, we can observe some interesting facts on the relationship between the values of betas and simulation results. As we can see above, when beta is -1, which means that each pixel is highly likely to have different signs with that of its neighboring nodes, the figure almost looks like having -1 and +1 alternatively for almost every pixel in the grid. Whereas, when beta's value is greater than 0 and gets increased (when beta is equal to 0.3 and 0.85), each pixel is likely to have same values with that of its neighboring ones and we can more clearly see there are bigger white parts and bigger black parts in the grid as beta gets increased. In following section, we will explore the method on estimating the beta value with the given observation Z.

3. **Statistical Inference on Beta in 2D Ising model (MRF) using MCMC algorithm**

The most fundamental problem we have in our estimation process stems from the normalizing constant in the model. We can intuitively think of using MCMC algorithm to estimate parameter $\beta$. However, given the observation Z, estimating beta requires the knowledge of normalizing constant $\psi(\beta)$ in the process of getting the acceptance ratio $\alpha(\beta, \beta')$. (i.e. both $\psi(\beta')$ and $\psi(\beta)$ cannot be cancelled out). Lei, et al (1999) introduced a brilliant way to solve this problem by utilizing the concept of pseudo-likelihood approximation (Besag, 1975) to the likelihood function we originally have.

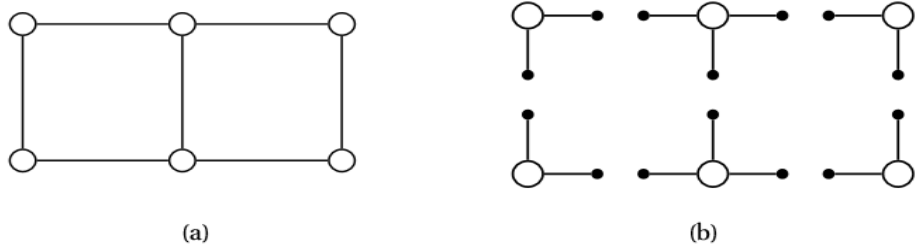

Figure 2. Pseudo-Likelihood in 2 x 3 Lattice

The PL approach is illustrated at Figure 2. for a 2d grid. We learn to predict each node, given all its neighbors. This objective is generally fast to compute since each full conditional density $P(Z_i|Z_{-i}, \beta)$ only requires summing over the states of a single node $Z_i$ in order to compute local normalization constant. Following is the specification of PL-approximation to our original likelihood function.

$$P(Z = z | \beta) \cong \prod_{i \in \Omega} P(Z = z_i | z_{-i}, \beta) = \prod_{i \in \Omega} P(Z = z_i | z_{N(i)}, \beta)$$

$$P(Z = z | \beta) \cong \prod_{i \in \Omega} \frac{\exp\{Z_i(\beta \sum_{j \in N(i)} Z_j)\}}{\sum_{Z_i \in \{-1,1\}} \exp\{Z_i(\beta \sum_{j \in N(i)} Z_j)\}}$$

Since our primary purpose is to estimate the parameter given the observation ($\hat{Z}$), the likelihood function can be written as follows:

$$P(\beta|Z = \hat{z}) \cong PL(\beta|Z = \hat{z}) = \prod_{i \in \Omega} \frac{\exp\{\hat{Z}_i(\beta \sum_{j \in N(i)} \hat{Z}_j)\}}{\sum_{Z_i \in \{-1,1\}} \exp\{Z_i(\beta \sum_{j \in N(i)} \hat{Z}_j)\}}$$

Note that, in process of calculating the denominator of each $P(Z = z_i | z_{-i}, \beta)$, provided that we know the values of given neighbors, only thing we need to do is to sum up the states of single node $Z_i$. Now, we can easily sample Markov Chain of $\beta$ from the approximated density, $PL(\beta|Z = \hat{z})$, using M-H algorithm.

At each time t, the next state $\beta^{t+1}$ is sampled from the proposal distribution, $g(\beta|\beta^t)$. This generated sequence of dependent random variables, $\{\beta^0, \beta^1, \beta^2, \beta^3 \ldots\}$, will gradually converge to the stationary distribution $\pi(\beta)$ after a sufficiently large number of iterations (n). If we consider a burn-in period, say m iterations, by the law of large number, we are able to estimate the value of true parameter $\beta$ by

$$\bar{\beta} = E(\beta|Z = \hat{z}) = \frac{1}{n-m} \sum_{i=m+1}^{n} \beta^i$$

This equation shows how the Markov chain of $\beta$ can be used to estimate $E(\beta|Z = \hat{z})$. Such a Markov Chain can be constructed by Metropolis-Hastings algorithm (Gilks, et al 1996). Let the proposal distribution $g(\beta'|\beta^t) = g(\beta^t|\beta')$ be the normal centered on the current value, $\beta^t$. Then, the acceptance probability, $\alpha(\beta^t, \beta')$, can be easily calculated as follows:

$$\alpha(\beta, \beta') = \min\left(1, \exp(log PL(\beta'|Z = \hat{z}) - log PL(\beta^t|Z = \hat{z}))\right)$$

$$= \min(1, \exp(\sum_{i \in \Omega} \hat{Z}_i \left(\beta' \sum_{j \in N(i)} \hat{Z}_j\right) - \sum_{i \in \Omega} \hat{Z}_i \left(\beta^t \sum_{j \in N(i)} \hat{Z}_j\right)$$

$$- \sum_{i \in \Omega} \log \sum_{Z_i \in \{-1,1\}} \exp\left\{Z_i \left(\beta' \sum_{j \in N(i)} \hat{Z}_j\right)\right\}$$

$$+ \sum_{i \in \Omega} \log \sum_{Z_i \in \{-1,1\}} \exp\left\{Z_i \left(\beta^t \sum_{j \in N(i)} \hat{Z}_j\right)\right\}))$$

If the acceptance ratio, $\alpha(\beta, \beta')$, is greater than or equal to the randomly generated number $u \sim \text{Unif}[0, 1]$, then we accept the newly proposed $\beta'$ as $\beta^{t+1}$. Otherwise, $\beta^{t+1}$ stays at previous beta, $\beta^t$. The Metropolis – Hastings algorithm can be summarized as following procedure.

| Parameter Estimation for MRF using Metropolis-Hastings Algorithm. |
|---|
| Initialize $\beta^0$; set t = 0 and T = Maximum number of Iteration |
| Start the Iteration |
| while t < T |
| Sample $\beta' \sim N(\beta^t, 1)$ |
| Generate a uniform random number: U ~ unif(0,1) |
| If U < $\alpha(\beta, \beta')$ then $\beta^{t+1} = \beta'$ |
| Else $\beta^{t+1} = \beta^t$ |
| t <- t + 1 |
| End the Iteration |

We performed several experiments to validate this methodology. On 125 x 125 lattice, we simulated Ising model with five different beta parameters: 1, 0.85, 0.3, -0.25, -1. The iteration numbers for the simulated data to be converged were somewhere around 2,000,000 for all five parameters. Since there is no definite convergence criteria, we had to check on our eyes whether behavior of Ising model changes or not as iteration number grows. In following graphs, we can check Markov Chain of $\beta$ converge within less than 300 iteration numbers when beta is equal to -1 (Left) and 1 (Right), respectively.

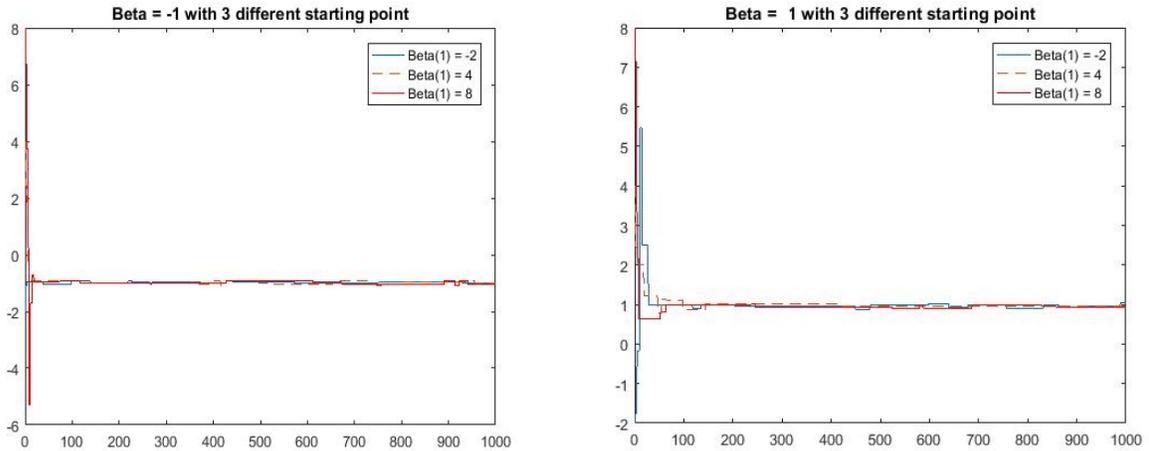

*Figure 3. Monte-Carlo output of Markov Chain { $\beta^0, \beta^1, \beta^2, \beta^3$ …. } when $\beta = -1, 1$*

In order to get acceptable parameters, the MCMC procedure described above should be repeated until stability of the Markov Chain is reached. The choice of starting points $\beta^0$, will not affect the stationary distribution if the chain is irreducible. Hence, here we start the algorithm with three different starting points: -2, 4, 8. The Markov chains converge in less than 300 iterations. The usual informal approach to detection of convergence is visual inspection of plots of the Monte-Carlo output { $\beta^0, \beta^1, \beta^2, \beta^3$ …. } (see Figure 3). Here we set burn-in period m = 500 times and n = 1000 times of total iterations. Table 1 displays the result of estimations for three different starting points with 5 different parameter settings.

Table 1. Estimation result of β in Markov Random Field using Pseudo-likelihood approximation

| Initial Point | Beta = -1 | Beta = -0.25 | Beta = 0.3 | Beta = 0.85 | Beta = 1 |
|---|---|---|---|---|---|
| -2 | -0.9837 | -0.2478 | 0.2952 | 0.7987 | 1.0096 |
| 4 | -0.9784 | -0.2473 | 0.3070 | 0.8241 | 1.0025 |
| 8 | -0.9816 | -0.2340 | 0.3044 | 0.8230 | 1.0196 |

## 4. Parameter estimation and image segmentation in HMRF with Spatial data.

Sometimes images are temporarily distorted due to various technical reasons. And true images(Z) are hidden under the noise data(Y). Here we study the methodology on how to estimate both the true image (Z) and parameters (θ) of the model at the same time. Here we use the term "Segmentation". Segmentation is a process which divides an image into several homogenous regions with smooth boundary. For smooth boundary, we use HMRF, which models spatial coherence between classes with MRF. For simplicity, we use the same Ising model as what we had defined in section 1:

$$P(Z = z \mid \beta) = \frac{1}{\psi(Z)} \exp\left\{\sum_{i \in \Omega} Z_i \left(\beta \sum_{j \in N(i)} Z_j\right)\right\} \quad (6)$$

where $\psi(Z)$ is a normalizing constant called the partition function, and each pixel at position i can take values either $Z_i = 1$ or $Z_i = -1$. Given Z, each component in Y is conditionally independent and has a distribution.

$$g(Y|Z = z, \mu_1, \mu_{-1}, \sigma^2) = \prod_{i \in \Omega} g(Y_i|Z_i = z_i, \mu_1, \mu_{-1}, \sigma^2)$$

Here we assume $g(\cdot \mid Z_i = z_i, \mu_1, \mu_{-1}, \sigma^2)$ to be the normal distribution $N(\mu, \sigma^2)$ with $\mu = \mu_1$ if $Z_i = 1$ and $\mu = \mu_{-1}$ if $Z_i = -1$. The neighborhood of an interior point i in Ω consists of four sites nearest to i in the 2D lattice Ω. We simulated the data from above model with $\beta = 0.2$, $\mu_1 = 2$, $\mu_{-1} = 0$, $\sigma^2 = 1$ in 125 x 125 lattice. Following figure plots a simulated dataset consisting of $\{Z_i \mid i \in \Omega\}$ and $\{Y_i \mid i \in \Omega\}$.

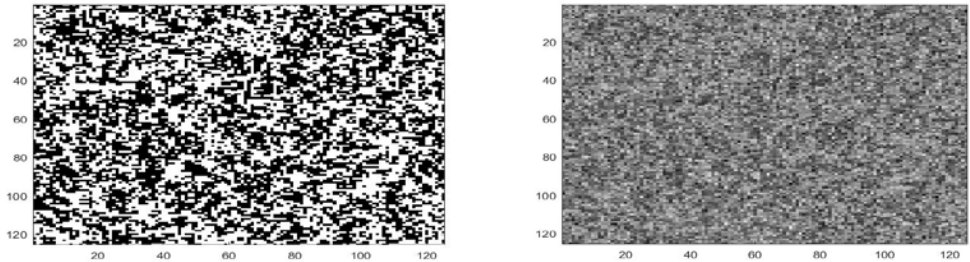

Figure 4. Simulated Z (left) and Y (Right) on 125 x 125 Lattice

It shows spatial dependency clearly in $\{Z_i \mid i \in \Omega\}$, but not in $\{Y_i \mid i \in \Omega\}$. With this simulated Y, we will now investigate the methodology on how to restore the original Z and parameter values of the model.

### 4.1. Overview of estimating procedure for the problem

We first need to observe how the likelihood function of our model looks like. Joint density function of $Y = \{Y_i \mid i \in \Omega\}$ and $Z = \{Z_i \mid i \in \Omega\}$ is as follows: For the simplicity of notation, we will denote $\theta = (\theta_1, \theta_2)$, where $\theta_1 = \beta$ and $\theta_2 = (\mu_1, \mu_{-1}, \sigma^2)$.

$$f_\theta(Z_1, \ldots, Z_\Omega, Y_1, \ldots, Y_\Omega) = P_{\theta_1}(Z_1, \ldots, Z_\Omega)\{\prod_{i \in \Omega} g_{\theta_2}(Y_i|Z_i)\}$$

Our primary purpose is to maximize the value of this likelihood function with respect to Z and θ. Since our model involves with latent variable, 'Z', we can intuitively think of using EM algorithm to obtain the MLE of the likelihood function. It consists of an expectation step (E-setp) and a maximization step (M-step) for each iteration. The E-step uses the current estimate $\theta^{(old)}$ to substitute for the conditional expectation of the complete-data log likelihood function: $Q(\theta|\theta^{(old)}) = E_{\theta^{(old)}}[\log f_\theta(Y, Z)|Y]$. M-step can be carried out by taking partial-derivatives of $Q(\theta|\theta^{(old)})$ with respect to each component of θ and solving the estimating equation $\dot{Q}(\theta|\theta^{(old)}) = 0$ to obtain the updated estimate of $\theta^{(new)}$. To compute the value of expected likelihood function $Q(\theta|\theta^{(old)})$ in E-step, we also need to estimate or restore the most likely state of Z, given Y and $\theta^{(old)}$. In other words, we use "*maximum a posteriori*" (MAP) to estimate for the true segmentation of image. The solution to our problem can be recapped as the iterations of following two steps:

---

MAP Estimation and EM algorithm

(Step 1). Given estimate of $\theta^{(old)}$, find the maximum a posteriori (MAP) estimate of Z, which is

$$\hat{Z}_{MAP} = argmax_Z P(Z|Y, \theta^{(old)})$$
$$= argmax_Z P_{\theta_1^{(Old)}}(Z_1, \ldots, Z_\Omega)\{\prod_{i \in \Omega} g_{\theta_2^{(Old)}}(Y_i|Z_i)\}$$

(Step 2). Using the estimate of Z, $\hat{Z}_{MAP}$, approximate the expected log-likelihood function $E_{\theta^{(old)}}[\log f_\theta(Y, Z)|Y]$ and update the estimates of θ as $\theta^{(new)}$. (EM)

---

## 4.2. MAP estimation of Z (Unobserved data) using Block Gibbs Sampler

Finding a MAP estimate in Step 1. can be computationally a very challenging task, and several Monte Carlo procedures have been proposed in statistical literature. As neatly pointed out by Lim et al (2007) 's paper, if we know the value of parameter $\theta$, the simplest procedure would be the iterative conditional mode (ICM), but since the likelihood of joint density is not a convex, it (ICM) is highly likely to be captured at local mode. Simulated Annealing had been suggested to resolve this local mode difficulty by choosing an appropriate tempering schedule. The most naïve procedure is, of course, based on samples from the posterior distribution, $P_\theta(Z|Y)$. Here, as we did in section 1, we can use Metropolis – Hastings algorithm to estimate Z with the given parameter $\theta$. But this single-site update requires lots of time for the data to be converged to the stationary state. It usually takes more than 1,000,000 times of iterations until the model to be converged. This procedure plays a pivotal role in determining the complexity of algorithm. Furthermore, sampling procedure is repeatedly used in estimating β with newly updated β for each iterations of EM. If these steps take long time to be completed, the entire time for the algorithm to be terminated should be exponentially increased. Block Gibbs Sampler provides a powerful tool to estimate the posterior distribution of unobserved data(Z) given the observed data (Y) in an HMRF context. As pointed out by Liu (2001, pp. 130 – 131), the Gibbs Sampler is a Markov Chain that converges geometrically to its stationary distribution, which is the posterior distribution of interest. The convergence rate of Z is dependent upon how $Z_i$ is correlated with each other. This fact leads Liu et al. (1994) to improve the efficiency of the Gibbs sampler by grouping highly correlated components to sample the blocks iteratively from their joint conditional distribution by the Block Gibbs Sampler.

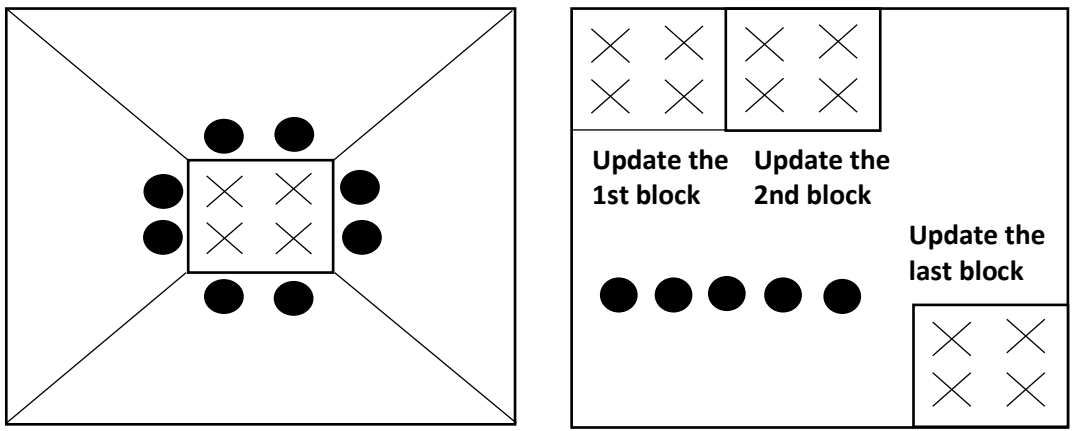

*Figure 5. Neighborhood system of 2x2 regular block (Left)*

*& how each block is updated without being overlapped with each other (Right)*

The key here is to define the block and joint conditional distribution. Block can be of any form, and we used the simplest 2 x 2 regular fixed block. Inner block's neighborhood can be set as 8 points which are connected by edges with each element in the block. Since Block Gibbs scheme is used not only for state estimation but also for parameter estimation, we split the case into two of its usage: Block Gibbs Sampler for calculating $E_{\theta^{(old)}}[\sum_{i\in\Omega} Z_i(\sum_{j\in N(i)} Z_j)|Y]$ and for calculating $E_{\beta}[\sum_{i\in\Omega} Z_i(\sum_{j\in N(i)} Z_j)]$. The one and only difference of these two procedures is on how to define the joint conditional distribution of each block. First, let us investigate the detailed procedure of Block Gibbs Sampler for posterior distribution.

### 4.2.1. Block Gibbs Sampler for posterior distribution

Let us denote the indexes of four nodes in ith block ($B_i$) as $Z_{B_i} = \{Z_{i1}, Z_{i2}, Z_{i3}, Z_{i4}\}$. By using conditional independence property of undirected graphical model, we can write a full conditional distribution of $Z_{B_i}$ as follows:

$$P_{\theta^{(old)}}(Z_{B_i} \mid \hat{Z}_{\setminus B_i}, Y_{B_i}) = P_{\theta^{(old)}}(Z_{B_i} \mid \hat{Z}_{N(B_i)}, Y_{B_i}) = \frac{q_{\theta^{(old)}}(Z_{B_i}, \hat{Z}_{N(B_i)}, Y_{B_i})}{\sum_{Z_{B_i} \in \{-1,+1\}^4} q_{\theta^{(old)}}(Z_{B_i}, \hat{Z}_{N(B_i)}, Y_{B_i})}$$

Here, $N(B_i)$ is a set of indexes of nodes surrounding the 2 x 2 regular block. The term at the nominator can be expressed as follows:

$$\exp\{\sum_{j=1}^{4} \frac{(Y_{ij} - \mu_{Z_{ij}}^{(old)})^2}{-2\sigma^2{}_{(old)}} + \beta^{(old)}(\sum_{(j,k)\in E_i} Z_{ij} Z_{ik} + \sum_{j=1}^{4} Z_{ij} \sum_{l \in L_{ij}} \hat{Z}_l)\}$$

where $E_i$ denotes a set of edges in $B_i$ and $L_{ij} = N(B_i) \cap N(Z_{ij})$. The first term considers total conditional dependencies between $Z_{B_i}$ and its corresponding observations $Y_{B_i}$. The second term takes both the interaction between nodes within the block and the interaction between the block's nodes and its neighborhood nodes into account. There are 16 possible combinations of $Z_{B_i}$, since each element in $Z_{B_i}$ can takes 2 values. Calculate the CDF value of this density, $P_{\theta^{(old)}}(Z_{B_i} \mid \hat{Z}_{\setminus B_i}, Y_{B_i})$, and pick the kth case out of 16, if CDF value corresponding to kth case first exceed the randomly generated number, $u \sim \text{unif}(0,1)$. As figure 5 indicates, update the next block in a way that there are no overlaps of elements between its neighboring blocks. Perform same update until all blocks in lattice are updated. We count this as a one iteration of Block Gibbs Sampler of posterior distribution. We performed 10,000 iterations and considers first 5,000 iterations as Burn-in period. With these remaining 5,000 lattice samples, we can calculate $E_{\theta^{(old)}}[\sum_{i\in\Omega} Z_i(\sum_{j\in N(i)} Z_j)|Y]$ and $E_{\theta^{(old)}}[\mathbb{I}(Z_i = 1)|Y]$ through Monte-Carlo approximation.

### 4.2.2. Block Gibbs Sampler for Gibbs distribution

In a process of estimating hyper-parameter, β, which will be introduced in a following section, we used Newton-Raphson method. Since the method requires gradient and hessian of the likelihood function, we need to calculate both $E_\beta[\sum_{i\in\Omega} Z_i(\sum_{j\in N(i)} Z_j)]$ and $var_\beta[\sum_{i\in\Omega} Z_i(\sum_{j\in N(i)} Z_j)]$ for every newly updated beta. We facilitated these calculations by using Block Gibbs Sampler. The only difference with that of conditional Block Gibbs is the form of full conditional distribution in a sense that we sample the samples from Gibbs distribution, not a posterior distribution. We define the full conditional distribution of Gibbs distribution (6) as follows:

$$P_\beta(Z_{B_i} \mid \hat{Z}_{\setminus B_i}) = P_\beta(Z_{B_i} \mid \hat{Z}_{N(B_i)}) = \frac{q_\beta(Z_{B_i}, \hat{Z}_{N(B_i)})}{\sum_{Z_{B_i} \in \{-1,+1\}^4} q_\beta(Z_{B_i}, \hat{Z}_{N(B_i)})}$$

Since we don't have to consider a conditional dependency between observed data and hidden data, the term at a nominator can be simply written as follows:

$$\exp\{\beta^{(old)}(\sum_{(j,k)\in E_i} Z_{ij}Z_{ik} + \sum_{j=1}^{4} Z_{ij} \sum_{l\in L_{ij}} \hat{Z}_l)\}$$

where $E_i$ denotes a set of edges in $B_i$ and $L_{ij} = N(B_i) \cap N(Z_{ij})$. Remaining sampling scheme is exactly same with that of posterior distribution. After obtaining enough lattice samples, we calculated the $E_\beta[\sum_{i\in\Omega} Z_i(\sum_{j\in N(i)} Z_j)]$ and $var_\beta[\sum_{i\in\Omega} Z_i(\sum_{j\in N(i)} Z_j)]$ through Monte-Carlo approximation. Since we are done with MAP approximation (Step 1), we are to elaborate how these approximated expectation values are used in EM algorithm (Step 2) to solve our parameter estimation problem.

### 4.3. Estimating parameters of Hidden Markov Random Field with Spatial data

The conditional expectation of complete data log-likelihood $Q(\theta|\theta^{(old)})$ can be written as follows:

$Q(\theta|\theta^{(old)})$
$= E_{\theta^{(old)}}\{\sum_{i\in\Omega} Z_i(\beta \sum_{j\in N(i)} Z_j) - \log \psi(\beta) \mid Y\}$
$+ E_{\theta^{(old)}}\{\mathbb{I}(Z_i = 1) \sum_{i\in\Omega}[\log g_{\theta_2}(Y_i|Z_i = 1) + \mathbb{I}(Z_i = -1)\log g_{\theta_2}(Y_i|Z_i = -1)] \mid Y\}$
$= \ell(\theta_1) + \ell(\theta_2)$ where $\theta_1 = \beta$ and $\theta_2 = (\mu_1, \mu_{-1}, \sigma^2)$

The estimation of $\theta_1, \theta_2$ can be separable in a sense that $\frac{\partial^2}{\partial \theta_1 \partial \theta_2} Q(\theta|\theta^{(old)}) = 0$.

Here we can get the conditional expectation of indicator functions as follows:

$$E_{\theta^{(old)}}[\mathbb{I}(Z_i = 1)|Y] = P_{\theta^{(old)}}(Z_i = 1|Y_i)$$

$$E_{\theta^{(old)}}[\mathbb{I}(Z_i = -1)|Y] = P_{\theta^{(old)}}(Z_i = -1|Y_i)$$

These values are calculated by conditional Block Gibbs Sampler of posterior distribution and used in process of estimating parameters of $\theta_2$. After taking expectation, likelihood functions of $\ell(\theta_1)$ and $\ell(\theta_2)$ are written as follows:

$$\ell(\theta_1) = \beta \, E_{\theta^{(old)}}\{\sum_{i \in \Omega} Z_i (\sum_{j \in N(i)} Z_j) | Y\} - \log \psi(\beta)$$

$$\ell(\theta_2) = \sum_{i \in \Omega} \{P_{\theta^{(old)}}(Z_i = 1|Y_i) \log g_{\theta_2}(Y_i|Z_i = 1) + P_{\theta^{(old)}}(Z_i = -1|Y_i) \log g_{\theta_2}(Y_i|Z_i = -1)\}$$

1) Estimation of $\mu_1$ and $\mu_{-1}$

   This problem is equivalent with the inference problem of the simplest Gaussian Mixture Model (GMM) with two mixed normal distributions. The estimation of parameter of $\mu_1$ is the solution to

$$\sum_{i \in \Omega} \left\{ P_{\theta^{(old)}}(Z_i = 1|Y_i) \frac{\partial \log g_{\theta_2}(Y_i|Z_i = 1)}{\partial \mu_1} + P_{\theta^{(old)}}(Z_i = -1|Y_i) \frac{\partial \log g_{\theta_2}(Y_i|Z_i = -1)}{\partial \mu_1} \right\} = 0$$

   Therefore, we got the solution as follows:

$$\mu_1 = \frac{\sum_{i \in \Omega} P_{\theta^{(old)}}(Z_i = 1|Y_i) * Y_i}{\sum_{i \in \Omega} P_{\theta^{(old)}}(Z_i = 1|Y_i)}$$

   Likewise, we also can get easily the case of $\mu_{-1}$.

$$\mu_{-1} = \frac{\sum_{i \in \Omega} P_{\theta^{(old)}}(Z_i = -1|Y_i) * Y_i}{\sum_{i \in \Omega} P_{\theta^{(old)}}(Z_i = -1|Y_i)}$$

2) Estimation of $\sigma^2$

   The estimation of $\sigma^2$ is the solution to the following equation: $\frac{\partial}{\partial \sigma^2} Q(\theta|\theta^{(old)}) = 0$, which is equivalent with:

$$\sum_{i \in \Omega} \left\{ P_{\theta^{(old)}}(Z_i = 1|Y_i) \frac{\partial \log g_{\theta_2}(Y_i|Z_i = 1)}{\partial \sigma^2} + P_{\theta^{(old)}}(Z_i = -1|Y_i) \frac{\partial \log g_{\theta_2}(Y_i|Z_i = -1)}{\partial \sigma^2} \right\} = 0$$

   After some tedious calculations, we can neatly write the solution of above equation in terms of $\sigma^2$ as follows:

$$\sigma^2 = \frac{1}{|\Omega|} \sum_{i \in \Omega} \{P_{\theta^{(old)}}(Z_i = 1|Y_i)(Y_i - \mu_1^{(old)})^2 + P_{\theta^{(old)}}(Z_i = -1|Y_i)(Y_i - \mu_{-1}^{(old)})^2\}$$

3) Estimation of hyper-parameter $\beta$

The estimate of hyper-parameter $\beta$ is the solution to: $\dot{\ell}(\theta_1) = 0$

$$\frac{\partial}{\partial \beta}[\beta\, E_{\theta^{(old)}}\{\sum_{i\in\Omega} Z_i(\sum_{j\in N(i)} Z_j)\,|Y\} - \log \psi(\beta)] = 0$$

Above equation can be rewritten as follows:

$$E_{\theta^{(old)}}\{\sum_{i\in\Omega} Z_i(\sum_{j\in N(i)} Z_j)\,|Y\} = \frac{\frac{\partial}{\partial \beta}\psi(\beta)}{\psi(\beta)}$$

where $\psi(\beta) = \sum_{Z\in\{-1,+1\}^{|\Omega|}} \exp\{\sum_{i\in\Omega} Z_i(\beta \sum_{j\in N(i)} Z_j)\}$, $Z = \{Z_i \mid i \in \Omega\}$.

Here,

$$\frac{\psi'(\beta)}{\psi(\beta)} = \sum_{Z\in\{-1,+1\}^{|\Omega|}} \frac{\sum_{i\in\Omega} Z_i(\sum_{j\in N(i)} Z_j) \exp\{\sum_{i\in\Omega} Z_i(\beta \sum_{j\in N(i)} Z_j)\}}{\sum_{Z\in\{-1,+1\}^{|\Omega|}} \exp\{\sum_{i\in\Omega} Z_i(\beta \sum_{j\in N(i)} Z_j)\}}$$

$$= E_\beta\left[\sum_{i\in\Omega} Z_i\left(\sum_{j\in N(i)} Z_j\right)\right]$$

The term $\frac{\psi'(\beta)}{\psi(\beta)}$ is an unconditional expectation with respect to Gibbs distribution in (6). So MLE of $\beta$ is the solution of following equation:

$$E_{\theta^{(old)}}\left[\sum_{i\in\Omega} Z_i\left(\sum_{j\in N(i)} Z_j\right)\,|Y\right] = E_\beta\left[\sum_{i\in\Omega} Z_i\left(\sum_{j\in N(i)} Z_j\right)\right]$$

LHS expectation can be approximated by Monte-Carlo sampling of posterior distribution, $P_{\theta^{(old)}}(Z = z\,|Y)$. RHS expectation can also be approximated by Monte-Carlo sampling of Gibbs Distribution in (1). LHS expectation is calculated once with respect to $\theta^{(old)}$ for every iteration of EM, and this should be considered as a constant at each updating procedure. Since above equation does not have a closed form solution, we have no option but to use recursive algorithm like Newton-Raphson or Bisection method. Here, we chose NR method over Bisection method, because it can be applied not only in Spatial HMRF model but also in Spatial-Temporal data in HMRF model.

Parameter updating scheme is extremely simple:

$$\beta^{(t)} = \beta^{(t-1)} - \frac{\ell'(\beta^{(t-1)})}{\ell''(\beta^{(t-1)})}$$

We can take double derivative with respect to $\beta$ on likelihood function $\ell(\theta_1)$.

$$\ell''(\beta) = \frac{\partial^2}{\partial^2 \beta}\{-\log \psi(\beta)\} = -\frac{\psi''(\beta)\psi(\beta) - (\psi'(\beta))^2}{(\psi(\beta))^2}$$

$$= -\left[\frac{\psi''(\beta)}{\psi(\beta)} - \left\{\frac{\psi'(\beta)}{\psi(\beta)}\right\}^2\right]$$

Then, we can immediately notice the term, $\frac{\psi'(\beta)}{\psi(\beta)}$, is $E_\beta[\sum_{i\in\Omega} Z_i(\sum_{j\in N(i)} Z_j)]$. First term can be calculated as follows:

$$\frac{\psi''(\beta)}{\psi(\beta)} = \sum_{Z\in\{-1,+1\}^{|\Omega|}} \frac{\{\sum_{i\in\Omega} Z_i(\sum_{j\in N(i)} Z_j)\}^2 \exp\{\sum_{i\in\Omega} Z_i(\beta \sum_{j\in N(i)} Z_j)\}}{\sum_{Z\in\{-1,+1\}^{|\Omega|}} \exp\{\sum_{i\in\Omega} Z_i(\beta \sum_{j\in N(i)} Z_j)\}}$$

$$= E_\beta\left[\left\{\sum_{i\in\Omega} Z_i\left(\sum_{j\in N(i)} Z_j\right)\right\}^2\right]$$

So, double derivative of likelihood function w.r.t. $\beta$ is equal to

$$\ell''(\beta) = -var_\beta\left[\left\{\sum_{i\in\Omega} Z_i\left(\sum_{j\in N(i)} Z_j\right)\right\}\right] \leq 0$$

Since $\ell''(\beta)$ is less than or equal to zero, $\ell(\theta_1)$ is a concave function and global optimal structure is guaranteed in our problem. We update the beta through the following calculation:

$$\beta^{(t)} = \beta^{(t-1)} + \frac{E_{\theta^{(old)}}\{\sum_{i\in\Omega} Z_i(\sum_{j\in N(i)} Z_j)|Y\} - E_{\beta^{(t-1)}}[\sum_{i\in\Omega} Z_i(\sum_{j\in N(i)} Z_j)]}{var_{\beta^{(t-1)}}[\{\sum_{i\in\Omega} Z_i(\sum_{j\in N(i)} Z_j)\}]}$$

We terminate the algorithm when $|\beta^{(t)} - \beta^{(t-1)}| < 10^{-3}$. If we have a good initial point to start the algorithm, we may only do one or two steps NR procedure. Following is the summary of NR method in estimating parameter $\beta$ in Hidden Markov Random Field. Note that $\theta^{(old)}$ is an already given set of parameters which are updated in each EM iteration.

> Newton-Raphson Method for estimating hyper-parameter, $\beta$.
> Given $E_{\theta^{(old)}}\{\sum_{i\in\Omega} Z_i(\sum_{j\in N(i)} Z_j) \mid Y\}$
> Start the Iteration
> while $|\beta^{(t)} - \beta^{(t-1)}| < 10^{-3}$
>
>     Calculate $E_{\beta^{(old)'}}[\sum_{i\in\Omega} Z_i(\sum_{j\in N(i)} Z_j)]$ & $var_{\beta^{(old)'}}[\{\sum_{i\in\Omega} Z_i(\sum_{j\in N(i)} Z_j)\}]$
>     by Block Gibbs Sampler of Gibbs Distribution.
> $$\beta^{(new)'} = \beta^{(old)'} - \frac{\ell'(\beta^{(old)'})}{\ell''(\beta^{(old)'})}$$
> End the Iteration
> Set $\beta^{(new)} = \beta^{(new)'}$

### 4.4. Estimation Result & Discussion

Table 2 gives the MLEs of $\theta = \{\mu_1, \mu_{-1}, \sigma^2, \beta\}$ for a single simulated dataset for $\Omega = \{1, \ldots, 48\} \times \{1, \ldots, 48\}$. To speed up the computation for the estimation, we reduced the size of lattice from 125 x 125 to 48 x 48. The convergence criteria used by the EM algorithm in section 3.3 is an upper bound of $10^{-3}$ for the size of each parameter increment. We performed several estimations with different parameter settings. The initial values of parameters were arbitrarily set and five estimations were executed in a range of [-1,0] and [0,1] respectively. The table reports the results of each estimated parameters. Each value in parenthesis (·,·,·,·) at first and seventh row of table denotes the true parameters of $\mu_1, \mu_{-1}, \sigma^2, \beta$ in a sequential order. The table also gives a CPU time (sec) carried out on a Laptop PC (Intel Core i7–5500 CPU(2.4GHz)). We found several interesting phenomena here through the results.

| 48 x 48 | (2, 1, 0.4, -0.2) | (2,1,0.4, -0.5) | (1.7, -1.5, 1.2, -0.7) | (0.7, -0.5, 1.6, -0.9) | (2,1,0.4, -1) |
|---|---|---|---|---|---|
| mu1 | 1.9925 | 1.9978 | 1.7232 | 0.7406 | 2.0199 |
| mu-1 | 1.0086 | 0.9774 | -1.5395 | -0.6244 | 1.0049 |
| sigma | 0.4087 | 0.4033 | 1.1714 | 1.5327 | 0.3949 |
| beta | -0.2144 | -0.4779 | -0.5811 | -0.5305 | -0.9789 |
| CPU time | 1112.63 | 316.25 | 803.85 | 903.55 | Not Terminate |
| 48 x 48 | (2,0,1,0.2) | (1.5, -1,1,0.5) | (2,1,0.4,0.6) | (2.3 -2 1.5 0.7) | (2.3 -2 1.5 1) |
| mu1 | 2.0006 | 1.5508 | 1.9729 | 2.3017 | 2.2875 |
| mu-1 | -0.0631 | -0.9986 | 1.0142 | -2.0692 | -2.2437 |
| sigma | 1.0060 | 0.9910 | 0.3965 | 1.4783 | 1.4936 |
| Beta | 0.2006 | 0.4850 | 0.5376 | 0.6042 | 0.6359 |
| CPU time | 617.26 | 369.08 | 398.97 | Not Terminate | 1049.45 |

*Table 2. Results of parameter estimations in HMRF with different parameter settings*

1. The EM algorithm works perfectly well when beta is set in a range of [-0.5, 0.5]. But outside of this region, the accuracy of beta estimation is reduced significantly. i.e. when beta is equal to -0.7 -0.9, 0.7, 1, NR poorly estimates beta values.

2. When true betas are set outside the range of [-0.5, 0.5] and initial values of betas are set within [-0.5, 0.5], poorly estimated beta values converge to the certain range of beta.

3. The estimations of $\theta_1 = \{\mu_1, \mu_{-1}, \sigma^2\}$ and $\theta_2 = \{\beta\}$ are independent in a sense that $\frac{\partial^2}{\partial \theta_1 \partial \theta_2} Q(\theta|\theta^{(old)}) = 0$. Check the table in cases where betas are poorly estimated. Algorithm estimates reasonably well all the parameters in $\theta_1$.

4. According to the initial values we set, sometimes, the algorithm doesn't terminate. i.e. In a case when θ = {2, 1, 0.4, -1}, if we set $\beta^{(initial)} = -0.9$.

Since partition function is computationally infeasible to get, it is impossible for us to draw the likelihood function with respect to beta. Instead we drew $\ell'(\beta)$ (score function of beta) to clarify why the abovementioned phenomena happen. Within the range of beta from -2 to 2 with 0.1 interval, values of $\ell'(\beta)$ for 41 betas are plotted in below figures. Four score functions are plotted when the true betas are 0.2, 0.6, 0.7, 1 respectively. The shape of all four functions looks somewhat similar to that of inverted sigmoid functions: whereas the functions' values sharply drops within the range of [-0.5, 0.5], it is formulated flat in range of [-2, -0.5] ∪ [0.5, 2]. This shape of functions can explain well all the three (1, 2, 4) weird behaviors of estimation process. If we start the algorithm with $\beta^{(initial)} \in [-0.5, 0.5]$, also if $\beta^{(true)} \in [-0.5, 0.5]$, then it will always succeed in estimating parameters precisely. But if $\beta^{(initial)} \in [-0.5, 0.5]$, and $\beta^{(true)} \notin [-0.5, 0.5]$, then the estimated beta will

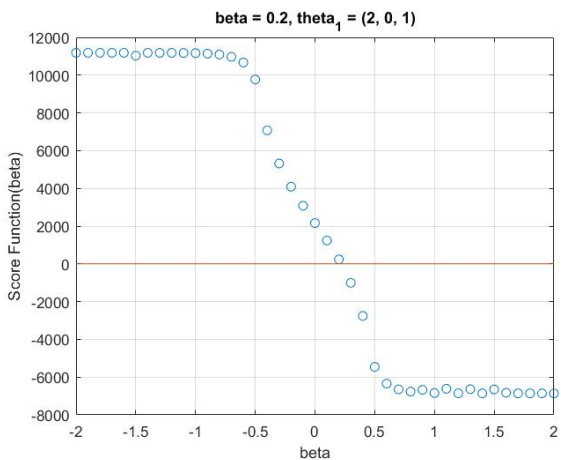
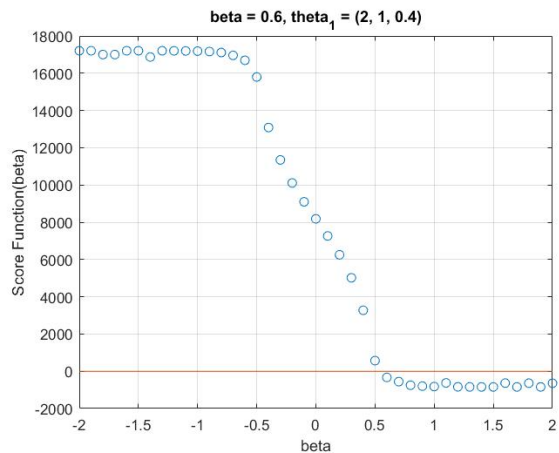

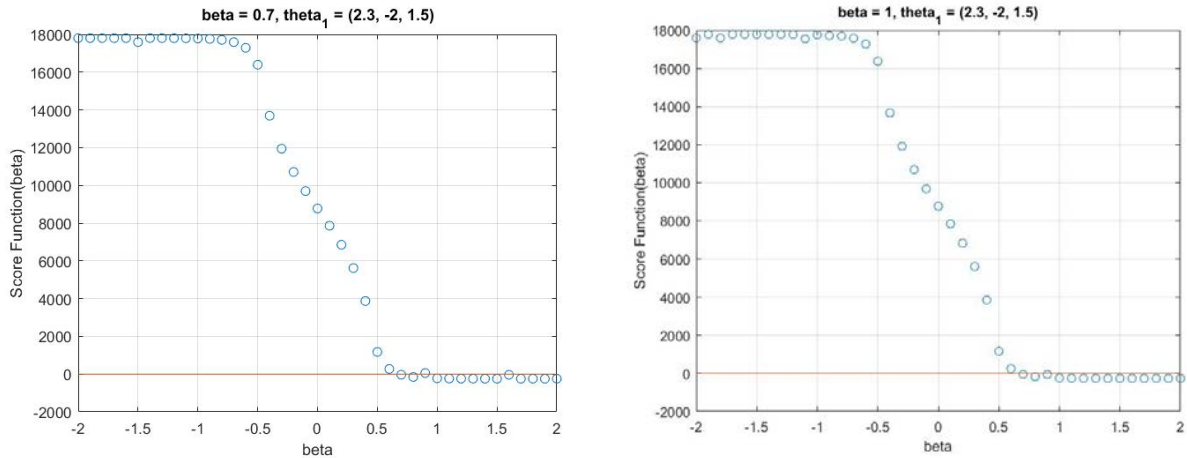

*Figure 5. See the points of betas where score function meets zero line*

be the one where it first hits the $\ell'(\beta) = 0$. As seen in the figures above and displayed at above table, when $\beta^{(true)} = 0.6, 0.7, 1$ then $\beta^{(estimated)} = 0.53, 0.60, 0.63$. In a case where $\beta^{(initial)} \notin [-0.5, 0.5]$, wherever $\beta^{(true)}$ lies, the algorithm never terminates, because Newton-Raphson method will diverge in a range of [-2, -0.5] ∪ [0.5, 2]. ($\because \ell''(\beta) \simeq 0$)

From the data point of view, when hidden states (Z) become sparse binary, in other words, when beta goes beyond certain threshold, most of the simulated data go to either -1 or +1. Therefore, a simulated dataset, itself, cannot contain much information on value of true beta. With this dataset, if we use one parameter Ising model as is in our case, likelihood function is formulated too flat nearby the true beta value. For example, in cases where the true beta is set as 0.6, 0.7, 0.9, 1.5, 2, it is highly likely for us to have sparse binary hidden state for all the cases. This leads us to an extremely hard situation in which we are not able to tell the differences between these cases from a finite number of data. (Figure 5.)

In recent studies, most researchers are too busy in utilizing HMRF without being conscious on theoretical aspects of HMRF. In most of the papers which are about statistical inferences on HMRF, researchers choose some "good" beta ($\beta^{(true)} \in [-0.5, 0.5]$), which produces reasonable amount of -1 and +1 in a dataset, and perform numerical studies. But in real data, most of the cases are not as good as what researchers are thinking. For instance, in genomic data, when "Differentially Expressed Gene" is sparse, then we have a problem in finding the network information of DEG. We are going to focus on addressing this sparse binary hidden state problem in our future research.

## 5. Parameter estimation of Hidden Markov Spatio-Temporal Random Field.

So far, we have focused on spatial data at a fixed time *t* and have assumed the signal $Z_t$ to be observable on fitting Markov random fields to these data. We now consider a more general setting of Hidden Markov Random Fields (HMRFs) of spatial-temporal data $Y_t$, $1 \leq t \leq T$, that are related to the signal $Z_t$ via the conditional densities $g_\theta(\cdot \mid \cdot)$ such that given Z = { $Z_{tw} : 1 \leq t \leq T, \omega \in \Omega$ }, the $Y_{tw}$ are conditionally independent with density function $g_\theta(Y_{tw}|Z_{tw})$. First order Markovian transition over time is assumed for $Z_t$ so that the joint density $h_\theta(\cdot)$ of $Z_t$ is of the following form (7).

$$h_\theta(Z) = \frac{1}{\psi(\theta)} \exp\left\{\sum_{t=1}^{T}\sum_{\omega\in\Omega} Z_{tw}\left(\beta \sum_{v\in N(\omega)} Z_{tv} + \alpha Z_{t-1,\omega}\right)\right\} \quad (7)$$

Here (7) basically augments the Ising model in the first paragraph of section 1 by adding the autoregressive term $\alpha Z_{t-1,\omega}$ at each voxel $\omega$ and time *t* for the logarithmic link function for the binary time series. Due to the assumption of Markovian transition over time, joint density can be expressed as the product of conditional densities $h_\theta(\cdot | Z_{t-1})$ as follows:

$$h_\theta(Z_1, Z_2, \ldots, Z_T) = h_\theta(Z_1)h_\theta(Z_2|Z_1)h_\theta(Z_3|Z_1,Z_2)\ldots h_\theta(Z_T|Z_1,Z_2,\ldots,Z_{T-1}) = \prod_{t=1}^{T} h_\theta(Z_t|Z_{t-1})$$

By using this relationship, likelihood function of observations over time t, $Y_t$ = { $Y_{tw}, \omega \in \Omega$ } has the form

$$\begin{aligned}
f_\theta(Y_1, Y_2, \ldots, Y_T) &= \sum_{Z_1, Z_2, \ldots, Z_T} f_\theta(Y_1, \ldots, Y_T, Z_1, \ldots, Z_T) \\
&= \sum_{Z_1, Z_2, \ldots, Z_T} g_{\theta_1}(Y_1, \ldots, Y_T \mid Z_1, \ldots, Z_T) h_{\theta_2}(Z_1, Z_2, \ldots, Z_T) \\
&= \sum_{Z_1, Z_2, \ldots, Z_T} \prod_{t=1}^{T}\left\{\prod_{\omega\in\Omega} g_{\theta_1}(Y_{tw}|Z_{tw})\right\} \prod_{t=1}^{T} h_{\theta_2}(Z_t|Z_{t-1}) \\
&= \sum_{Z_1, Z_2, \ldots, Z_T} \prod_{t=1}^{T}\left\{\left[\prod_{\omega\in\Omega} g_{\theta_1}(Y_{tw}|Z_{tw})\right] h_{\theta_2}(Z_t|Z_{t-1})\right\}
\end{aligned}$$

in which $h_{\theta_2}(Z_t|Z_{t-1}) = h_{\theta_2}(Z_1)$ for the case *t* = 1. Here, since $g_\theta$ only depends on a sub-vector $\theta_1$ of $\theta$ and $h_\theta$ only depends on a sub-vector $\theta_2$ of $\theta$, we differentiate the notation of two densities. We also assume here the model as Gaussian Hidden Markov Spatial-Temporal random field, where the conditional density $g_\theta(\cdot|Z_{tw})$ is the normal distribution $N(\mu, \sigma^2)$ with $\mu = \mu_1$ if $Z_i = 1$ and $\mu = \mu_{-1}$ if $Z_i = -1$. (i.e. $\theta_1 = (\mu_1, \mu_{-1}, \sigma^2)$, $\theta_2 = (\alpha, \beta)$) Since the estimation procedures are exactly same with that of what we had done at section 3, we will be going to provide a much simpler explanation on parameter estimation processes for HMRFs for spatial-temporal data.

We kick off the problem by getting a conditional expectation of complete log-likelihood function. First, we define a complete log-likelihood function.

$$\log f_\theta(Y_1, \ldots, Y_T, Z_1, \ldots, Z_T) = \log \prod_{t=1}^{T} \left\{ \left[ \prod_{\omega \in \Omega} g_{\theta_1}(Y_{tw}|Z_{tw}) \right] h_{\theta_2}(Z_t|Z_{t-1}) \right\}$$

$$= \log \prod_{t=1}^{T} \prod_{\omega \in \Omega} g_{\theta_1}(Y_{tw}|Z_{tw}) + \log h_{\theta_2}(Z_1, Z_2, \ldots, Z_T)$$

$$= \sum_{t=1}^{T} \sum_{\omega \in \Omega} \{\mathbb{I}(Z_{tw} = 1)\log g_{\theta_1}(Y_{tw}|Z_{tw} = 1) + \mathbb{I}(Z_{tw} = -1)\log g_{\theta_1}(Y_{tw}|Z_{tw} = -1)\}$$

$$+ \beta \sum_{t=1}^{T} \sum_{\omega \in \Omega} Z_{tw} \sum_{v \in N(\omega)} Z_{tv} + \alpha \sum_{t=1}^{T} \sum_{\omega \in \Omega} Z_{tw} Z_{t-1,\omega} - \log \psi(\theta_2)$$

where $\psi(\theta_2) = \sum_{Z_1, \ldots, Z_T \in \{-1,1\}^{T|\Omega|}} \exp\{\sum_{t=1}^{T} \sum_{\omega \in \Omega} Z_{tw}(\beta \sum_{v \in N(\omega)} Z_{tv} + \alpha Z_{t-1,\omega})\}$

For simplicity, let us put

$$T_1(Z) = \sum_{t=1}^{T} \sum_{\omega \in \Omega} Z_{tw} \sum_{v \in N(\omega)} Z_{tv}$$

$$T_2(Z) = \sum_{t=1}^{T} \sum_{\omega \in \Omega} Z_{tw} Z_{t-1,\omega}$$

Then conditional expectation of complete data log-likelihood function can be expressed as follows: $Q(\theta|\theta^{(old)}) = E_{\theta^{(old)}}[\log f_\theta(Y, Z)|Y]$

$$Q(\theta|\theta^{(old)}) = \ell(\theta_1) + \ell(\theta_2) \text{ where } \theta_1 = (\mu_1, \mu_{-1}, \sigma^2) \text{ and } \theta_2 = (\alpha, \beta)$$

$$\ell(\theta_1|\theta^{(old)}) = \sum_{t=1}^{T} \sum_{\omega \in \Omega} \{P_{\theta^{(old)}}(Z_{tw} = 1|Y_{tw})\log g_{\theta_1}(Y_{tw}|Z_{tw} = 1) + P_{\theta^{(old)}}(Z_{tw} = -1|Y_{tw})\log g_{\theta_1}(Y_{tw}|Z_{tw} = -1)\}$$

$$\ell(\theta_2|\theta^{(old)}) = \beta E_{\theta^{(old)}}\{T_1(Z)|Y\} + \alpha E_{\theta^{(old)}}\{T_2(Z)|Y\} - \log \psi(\theta_2)$$

Here, also we can independently estimate the parameters of $\theta_1$ and $\theta_2$, in a sense that $\frac{\partial^2}{\partial \theta_1 \partial \theta_2} Q(\theta|\theta^{(old)}) = 0$.

1) Estimation of $\mu_1$ and $\mu_{-1}$

   Equivalent with the case above at sub-section 3.3's first estimation procedure, we can easily get the MLE of $\mu_1$ and $\mu_{-1}$ as the solutions of following equations.

   $$\sum_{t=1}^{T}\sum_{w\in\Omega}\left\{P_{\theta^{(old)}}(Z_{tw}=1|Y_{tw})\frac{\partial\,log g_{\theta_1}(Y_{tw}|Z_{tw}=1)}{\partial\,\mu_1}\right.$$
   $$\left.+P_{\theta^{(old)}}(Z_{tw}=-1|Y_{tw})\frac{\partial log g_{\theta_1}(Y_{tw}|Z_{tw}=-1)}{\partial\,\mu_1}\right\}=0$$

   Therefore, we got the solution as follows:

   $$\mu_1=\frac{\sum_{t=1}^{T}\sum_{w\in\Omega}P_{\theta^{(old)}}(Z_{tw}=1|Y_{tw})*Y_{tw}}{\sum_{t=1}^{T}\sum_{w\in\Omega}P_{\theta^{(old)}}(Z_{tw}=1|Y_{tw})}$$

   Likewise, we also can get easily the case of $\mu_{-1}$.

   $$\mu_{-1}=\frac{\sum_{t=1}^{T}\sum_{w\in\Omega}P_{\theta^{(old)}}(Z_{tw}=-1|Y_{tw})*Y_{tw}}{\sum_{t=1}^{T}\sum_{w\in\Omega}P_{\theta^{(old)}}(Z_{tw}=-1|Y_{tw})}$$

2) Estimation of $\sigma^2$

   The estimation of $\sigma^2$ is the solution to the following equation: $\frac{\partial}{\partial\,\sigma^2}Q(\theta|\theta^{(old)})=0$, which is equivalent with:

   $$\sum_{t=1}^{T}\sum_{w\in\Omega}\left\{P_{\theta^{(old)}}(Z_{tw}=1|Y_{tw})\frac{\partial\,log g_{\theta_1}(Y_{tw}|Z_{tw}=1)}{\partial\,\sigma^2}\right.$$
   $$\left.+P_{\theta^{(old)}}(Z_{tw}=-1|Y_{tw})\frac{\partial log g_{\theta_1}(Y_{tw}|Z_{tw}=-1)}{\partial\,\sigma^2}\right\}=0$$

   After some tedious calculations, we can neatly write the solution of above equation in terms of $\sigma^2$ as follows:

   $$\sigma^2=\frac{1}{T|\Omega|}\sum_{t=1}^{T}\sum_{w\in\Omega}\{P_{\theta^{(old)}}(Z_{tw}=1|Y_{tw})(Y_{tw}-\mu_1^{(old)})^2+P_{\theta^{(old)}}(Z_{tw}=-1|Y_{tw})(Y_{tw}-\mu_{-1}^{(old)})^2\}$$

3) Estimation of hyper-parameter β and α

   We also apply Newton-Raphson Method for the estimation of two parameters β and α. NR step for two parameters can be defined as follows:

   $$\begin{bmatrix}\hat{\beta}^{(t)}\\\hat{\alpha}^{(t)}\end{bmatrix}=\begin{bmatrix}\hat{\beta}^{(t-1)}\\\hat{\alpha}^{(t-1)}\end{bmatrix}-H(\hat{\beta}^{(t-1)},\hat{\alpha}^{(t-1)})^{-1}\begin{bmatrix}\frac{\partial}{\partial\beta}\ell(\theta_2|\theta^{(t-1)})\\\frac{\partial}{\partial\alpha}\ell(\theta_2|\theta^{(t-1)})\end{bmatrix}$$

   $$where\;H(\hat{\beta}^{(t-1)},\hat{\alpha}^{(t-1)})=\begin{pmatrix}\frac{\partial^2}{\partial^2\beta}\ell(\theta_2|\theta^{(t-1)}) & \frac{\partial^2}{\partial\beta\partial\alpha}\ell(\theta_2|\theta^{(t-1)})\\\frac{\partial^2}{\partial\beta\partial\alpha}\ell(\theta_2|\theta^{(t-1)}) & \frac{\partial^2}{\partial^2\alpha}\ell(\theta_2|\theta^{(t-1)})\end{pmatrix}$$

3 - 1) Computation of Gradient

$$\frac{\partial}{\partial \beta} \ell(\theta_2 | \theta^{(t-1)}) = E_{\theta^{(old)}}\{T_1(Z)|Y\} - \frac{\frac{\partial}{\partial \beta} \psi(\theta_2^{(t-1)})}{\psi(\theta_2^{(t-1)})}$$

$$\frac{\partial}{\partial \alpha} \ell(\theta_2 | \theta^{(t-1)}) = E_{\theta^{(old)}}\{T_2(Z)|Y\} - \frac{\frac{\partial}{\partial \alpha} \psi(\theta_2^{(t-1)})}{\psi(\theta_2^{(t-1)})}$$

where,

$$\frac{\frac{\partial}{\partial \beta} \psi(\theta_2^{(t-1)})}{\psi(\theta_2^{(t-1)})} = \sum_{Z_1,\ldots,Z_T \in \{-1,1\}^{T|\Omega|}} \frac{T_1(Z) \exp\{\beta^{(t-1)} T_1(Z) + \alpha^{(t-1)} T_2(Z)\}}{\sum_{Z_1,\ldots,Z_T \in \{-1,1\}^{T|\Omega|}} \exp\{\beta^{(t-1)} T_1(Z) + \alpha^{(t-1)} T_2(Z)\}} = E_{\theta_2^{(t-1)}}[T_1(Z)]$$

$$\frac{\frac{\partial}{\partial \alpha} \psi(\theta_2^{(t-1)})}{\psi(\theta_2^{(t-1)})} = \sum_{Z_1,\ldots,Z_T \in \{-1,1\}^{T|\Omega|}} \frac{T_2(Z) \exp\{\beta^{(t-1)} T_1(Z) + \alpha^{(t-1)} T_2(Z)\}}{\sum_{Z_1,\ldots,Z_T \in \{-1,1\}^{T|\Omega|}} \exp\{\beta^{(t-1)} T_1(Z) + \alpha^{(t-1)} T_2(Z)\}} = E_{\theta_2^{(t-1)}}[T_2(Z)]$$

Above two expectation values can be approximated through Monte Carlo samples from Gibbs distribution (2), generated at parameter setting. $\theta_2^{(t-1)} = (\alpha^{(t-1)}, \beta^{(t-1)})$. Here, we use Block gibbs sampler scheme to efficiently sample from Gibbs distribution.

3 – 2) Computation of Hessian
Each element in Hessian can be calculated in similar fashion as follows:

$$\frac{\partial^2}{\partial^2 \beta} \ell(\theta_2 | \theta^{(t-1)})$$

$$= \frac{\partial^2}{\partial^2 \beta}\{-\log \psi(\theta^{(t-1)})\} = -\left[\frac{1}{\psi(\theta_2^{(t-1)})} * \frac{\partial^2 \psi(\theta_2^{(t-1)})}{\partial^2 \beta} - \left\{\frac{\frac{\partial}{\partial \beta} \psi(\theta_2^{(t-1)})}{\psi(\theta_2^{(t-1)})}\right\}^2\right]$$

$$= -\left[E_{\theta_2^{(t-1)}}[T_1(Z)^2] - \{E_{\theta_2^{(t-1)}}[T_1(Z)]\}^2\right] = -var_{\theta_2^{(t-1)}}(T_1(Z))$$

where the term $\frac{1}{\psi(\theta_2^{(t-1)})} * \frac{\partial^2 \psi(\theta_2^{(t-1)})}{\partial^2 \beta}$ can be calculated as follows :

$$\frac{\frac{\partial^2 \psi(\theta_2^{(t-1)})}{\partial^2 \beta}}{\psi(\theta_2^{(t-1)})} = \sum_{Z_1,\ldots,Z_T \in \{-1,1\}^{T|\Omega|}} \frac{T_1(Z)^2 \exp\{\beta^{(t-1)} T_1(Z) + \alpha^{(t-1)} T_2(Z)\}}{\sum_{Z_1,\ldots,Z_T \in \{-1,1\}^{T|\Omega|}} \exp\{\beta^{(t-1)} T_1(Z) + \alpha^{(t-1)} T_2(Z)\}} = E_{\theta_2^{(t-1)}}[T_1(Z)^2]$$

Similarly, $\frac{\partial^2}{\partial^2 \alpha} \ell(\theta_2|\theta^{(t-1)})$ can be computed exactly same with $\frac{\partial^2}{\partial^2 \beta} \ell(\theta_2|\theta^{(t-1)})$.

$$\frac{\partial^2}{\partial^2 \beta} \ell(\theta_2|\theta^{(t-1)}) = -var_{\theta_2^{(t-1)}}(T_2(Z))$$

Lastly,

$$\frac{\partial^2}{\partial \alpha \partial \beta} \ell(\theta_2|\theta^{(t-1)})$$

$$= \frac{\partial^2}{\partial \alpha \partial \beta} \{-\log \psi(\theta^{(t-1)})\}$$

$$= -\left[\frac{1}{\psi(\theta_2^{(t-1)})^2} * \left\{\frac{\partial^2 \psi(\theta_2^{(t-1)})}{\partial \alpha \, \partial \beta} - \frac{\partial \psi(\theta_2^{(t-1)})}{\partial \alpha} \frac{\partial \psi(\theta_2^{(t-1)})}{\partial \beta}\right\}\right]$$

$$= -\left[E_{\theta_2^{(t-1)}}[T_1(Z) T_2(Z)] - E_{\theta_2^{(t-1)}}[T_1(Z)] * E_{\theta_2^{(t-1)}}[T_2(Z)]\right]$$

$$= -cov_{\theta_2^{(t-1)}}(T_1(Z), T_2(Z))$$

The Hessian of the $\ell(\theta_2|\theta^{(t-1)})$ can be defined in following form:

$$H(\hat{\beta}^{(t-1)}, \hat{\alpha}^{(t-1)}) = \begin{pmatrix} \frac{\partial^2}{\partial^2 \beta} \ell(\theta_2|\theta^{(t-1)}) & \frac{\partial^2}{\partial \beta \partial \alpha} \ell(\theta_2|\theta^{(t-1)}) \\ \frac{\partial^2}{\partial \beta \partial \alpha} \ell(\theta_2|\theta^{(t-1)}) & \frac{\partial^2}{\partial^2 \alpha} \ell(\theta_2|\theta^{(t-1)}) \end{pmatrix}$$

$$= -\begin{pmatrix} var_{\theta_2^{(t-1)}}(T_1(Z)) & cov_{\theta_2^{(t-1)}}(T_1(Z), T_2(Z)) \\ cov_{\theta_2^{(t-1)}}(T_1(Z), T_2(Z)) & var_{\theta_2^{(t-1)}}(T_2(Z)) \end{pmatrix}$$

3 – 3) Stopping criteria

We stop the algorithm when the $l_2$ norm of the vector, which is a difference between newly updated parameters and previous parameters before being updated, is less than or equal to $\epsilon_{NR}$. In our estimation, we set $\epsilon_{NR}$ as $10^{-3}$.

$$\sqrt{(\hat{\beta}^{(t)} - \hat{\beta}^{(t-1)})^2 + (\hat{\alpha}^{(t)} - \hat{\alpha}^{(t-1)})^2} \leq \epsilon_{NR}$$

## 5.1. Block Gibbs Sampler for posterior distribution (Spatio – Temporal)

Let us denote the indexes of four nodes in ith block at time t as $Z_{B_{it}} = \{Z_{i1t}, Z_{i2t}, Z_{i3t}, Z_{i4t}\}$. By using conditional independence property of undirected graphical model, we can write a full conditional distribution of $Z_{B_{it}}$ with respect to posterior distribution of observations $Y_{B_{it}}$ as follows:

$$P_{\theta^{(old)}}(Z_{B_{it}} \mid \hat{Z}_{\backslash B_{it}}, \hat{Y}_{B_{it}}) = P_{\theta^{(old)}}(Z_{B_{it}} \mid \hat{Z}_{N(B_{it})}, \hat{Y}_{B_{it}}) = \frac{q_{\theta^{(old)}}(Z_{B_{it}}, \hat{Z}_{N(B_{it})}, \hat{Y}_{B_{it}})}{\sum_{Z_{B_{it}} \in \{-1,+1\}^4} q_{\theta^{(old)}}(Z_{B_{it}}, \hat{Z}_{N(B_{it})}, \hat{Y}_{B_{it}})}$$

Here, $N(B_{it})$ is a set of indexes of nodes surrounding the 2 x 2 regular block. The term at the nominator can be expressed as follows:

$$\exp\left\{ \sum_{j=1}^{4} \frac{(\hat{Y}_{ijt} - \mu_{Z_{ijt}}^{(old)})^2}{-2\sigma^2_{(old)}} + \beta^{(old)} \left( \sum_{(i,j,t) \sim (i,k,t)} Z_{ijt} Z_{ikt} + \sum_{j=1}^{4} Z_{ijt} \sum_{l \in L_{ijt}} \hat{Z}_l \right) + \alpha^{(old)} \sum_{j=1}^{4} Z_{ijt} (\hat{Z}_{ij,t-1} + \hat{Z}_{ij,t+1}) \right\}$$

where the symbol $(i,j,t) \sim (i,k,t)$ indicates that nodes $(i,j,t)$ $and$ $(i,k,t)$ are neighbors within the block $B_{it}$, and $L_{ijt} = N(B_{it}) \cap N(Z_{ijt})$ denotes an intersection of a set of indexes of neighboring nodes of block $B_{it}$ and a set of indexes of neighboring nodes of $Z_{ijt}$. Note that the only difference of the form of full conditional distribution for block gibbs between spatial data and spatial-temporal data is that the absence or presence of the first order auto-regressive term. Since this is exactly same case with that of block gibbs sampler for spatial-temporal gibbs distribution, we will skip this part and move onto the result and discussion of our simulation result on spatial-temporal HMRF.

| 24 x 24 x 40 | (0.3, 1.5, 0.49, 0.3, 0.1) | 48 x 48 x 90 | (0, 2, 1, 0.1, 0.1) |
|---|---|---|---|
| mu1 | 0.31496 | mu1 | -0.00982 |
| mu-1 | 1.52375 | mu-1 | 2.00422 |
| sigma | 0.48821 | sigma | 0.98416 |
| beta | 0.30005 | beta | 0.10278 |
| alpha | 0.09969 | alpha | 0.09367 |
| CPU time (min) | 141.5 | CPU time | 275.0 |

*Table 3. Results of parameter estimations in Spatial-Temporal HMRF*

### 5.2. Estimation result for simulation study of Spatial-Temporal HMRF

We performed two simulation studies for the lattice size 24 x 24 x 90 and 48 x 48 x 90 respectively. Table 3 gives the MLE of θ = { $\mu_1$, $\mu_{-1}$, $\sigma^2$, β, α} for a single simulated dataset. Here, we also generated each dataset by using 2 x 2 block gibbs sampling method. Since the heavy computational load of our algorithm, it was coded in C, not in MATLAB. The table also gives a CPU time (min) carried out on a Laptop PC (Intel Core i7 – 5500 CPU (2.4 GHz)). As the table displays, although the algorithm accurately estimates all the parameters, the time it takes to complete the estimation is quite long. Furthermore, though not displayed in the table, if either the parameter β or α goes beyond certain threshold, it will also produce spare binary state data, and we have a trouble in estimating both parameters. As for improving the speed of parameter estimation of spatial-temporal HMRF, Dr. Tze. L. Lai and Dr. Lim (2015) suggested a new solution of using block likelihood function by partitioning the whole dataset into smaller ones and compute the estimates in parallel. This method significantly reduces the estimation time and provides us with new insights on our next research directions.

# References

1. Siddhartha Chib; Edward Greenberg, (1995): Understanding the Metropolis-Hastings Algorithm, American Statistical Association

2. Lei W, Jun Liu, et al. (1999): MRF parameter estimation by MCMC method, Pattern recognition

3. Martin J. Wainwright, Michael Jordan (2008): Graphical Models, Exponential Families, and Variational Inference, Foundation and Trends in Machine Learning.

4. Tze Leung Lai, Johan Lim (2015) : Asymptotically Efficient Parameter Estimation in Hidden Markov Spatio-Temporal Random Fields, Statistica Sincia.

5. Johan Lim, Kyungsuk Pyun, Chee Sun Won (2007): Image Segmentation Using Hidden Markov Gauss Mixture Models, IEEE

6. Besag, Bartolucci (2002) : A recursive algorithm for Markov Random Fields, Biometrika

7. Ruslan : Learning in Markov Random Fields using Tempered Transitions

8. Zhou, Leahy, Qi (1997) Approximate Maximum Likelihood Hyperparameter Estimation for Gibbs Priors. IEEE

9. N. Friel, A.N. Pettitt et al : Bayesian inference in hidden Markov Random fields for binary data defined on large lattice.

10. Johan Lim : Review of Estimation of hidden Markov Random Field